\documentclass{article}

\usepackage{microtype}
\usepackage{graphicx}
\usepackage{subfigure}
\usepackage{booktabs} 
\usepackage{tabularx}
\usepackage{listings} 
\usepackage{xcolor}
\usepackage{color}
\usepackage[T1]{fontenc}
\usepackage{multirow}
\usepackage{amsmath}
\usepackage{lipsum}
\usepackage{multicol}
\usepackage{threeparttable}

\usepackage{hyperref}

\def\ie{\emph{i.e.}}
\def\eg{\emph{e.g.}}

\def\etc{\emph{etc}}
\newcommand\bb[1]{\textbf{#1}}

\newcommand\U[1]{\underline{#1}}

\usepackage[accepted]{sysml2020}

\sysmltitlerunning{MNN: A Universal and Efficient Inference Engine}

\begin{document}

\twocolumn[
\sysmltitle{MNN: A Universal and Efficient Inference Engine}




\begin{sysmlauthorlist}
    \sysmlauthor{Xiaotang Jiang}{ali}
    \sysmlauthor{Huan Wang}{neu}
    \sysmlauthor{Yiliu Chen}{ali}
    \sysmlauthor{Ziqi Wu}{ali}
    \sysmlauthor{Lichuan Wang}{ali}
    \sysmlauthor{Bin Zou}{ali}
    \sysmlauthor{Yafeng Yang}{ali}
    \sysmlauthor{Zongyang Cui}{ali}
    \sysmlauthor{Yu Cai}{ali}
    \sysmlauthor{Tianhang Yu}{ali}
    \sysmlauthor{Chengfei Lv}{ali}
    \sysmlauthor{Zhihua Wu}{ali}
\end{sysmlauthorlist}

\sysmlaffiliation{ali}{Alibaba Group, Hangzhou, China.}
\sysmlaffiliation{neu}{Department of Electrical and Computer Engineering, Notheastern University, Boston, USA}
\sysmlcorrespondingauthor{Ziqi Wu}{mingyi.wzq@alibaba-inc.com}

\sysmlkeywords{MNN, Mobile Inference Engine}
\vskip 0.3in

\begin{abstract}
Deploying deep learning models on mobile devices draws more and more attention recently. However, designing an efficient inference engine on devices is under the great challenges of model compatibility, device diversity, and resource limitation. To deal with these challenges, we propose Mobile Neural Network (MNN), a universal and efficient inference engine tailored to mobile applications. In this paper, the contributions of MNN include:
(1) presenting a mechanism called pre-inference that manages to conduct runtime optimization;
(2) delivering thorough kernel optimization on operators to achieve optimal computation performance;
(3) introducing backend abstraction module which enables hybrid scheduling and keeps the engine lightweight.
Extensive benchmark experiments demonstrate that MNN performs favorably against other popular lightweight deep learning frameworks. MNN is available to public at: \href{https://github.com/alibaba/MNN}{https://github.com/alibaba/MNN}.
\end{abstract} 
]



\printAffiliationsAndNotice{}  

\section{Introduction}
\label{intro}
Deep learning has become the de facto method for artificial intelligence in various tasks including computer vision, user intention recognition~\cite{guo2019buying}, and autopilot~\cite{LecBenHin15}. As edge devices (\eg, smartphones, IoT devices, wearable devices) are ubiquitous now,  deep learning on edge devices, especially mobile devices, attracts growing attention~\cite{shi2016edge}. There are many advantages for deep learning on mobiles, for example, low latency, privacy protection, and personalized service. To make full use of on-device deep learning technology, inference engines tailored to mobile devices have been developed and extensively used in mobile applications, for example, TF-Lite~\cite{tflite}\cite{tflite}, NCNN~\cite{ncnn}, CoreML~\cite{coreml}, \etc.

The major challenges for mobile inference engines can be categorized into three aspects: model compatibility, device diversity, and resource limitation.

\begin{figure}[t]
    \centering
    \includegraphics[width=\linewidth]{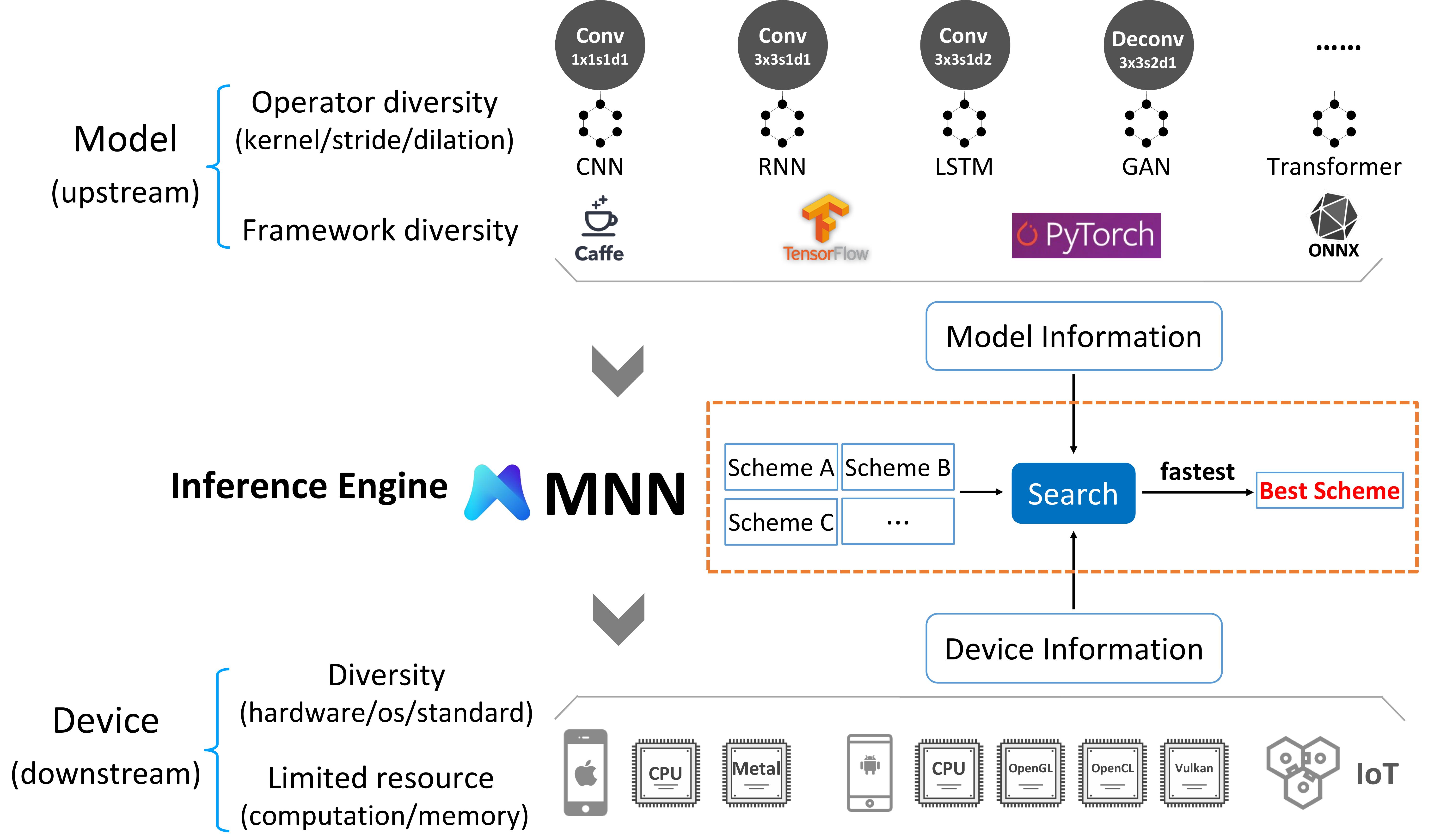}
    \caption{Overview of the proposed Mobile Neural Network.}
    \label{fig:overview}
\end{figure}

\textbf{(1) Model compatibility}. Most deep learning models deployed on mobile devices are trained from well-known deep learning frameworks such as TensorFlow~\cite{abadi2016tensorflow}, PyTorch~\cite{pytorch}, Caffe~\cite{jia2014caffe}, CNTK~\cite{cntk}, MXNet~\cite{chen2015mxnet}. As a result, it is a basic requirement that an inference engine should have the model compatibility for different formats and different operators. More importantly, the engine should also allow for proper scalability to support new operators emerging in the future.

\textbf{(2) Device diversity}. Almost every well-known mobile applications are used extensively on various devices, ranging from low-end equipment with single-core CPU to high-end equipment with co-processors such as Apple Neural Engine (ANE). To achieve great performance on various devices, mobile inference engines have to take hardware architectures or even device vendors (like ARM Mali GPU or Qualcomm Adreno GPU) into consideration. In addition, the engine is also expected to take care of the software diversity problem well, such as different operating systems (Android/iOS/embedded OS) and different solution standards (OpenCL~\cite{opencl}/OpenGL~\cite{opengl}/Vulkan~\cite{vulkan} for Android GPU).

\textbf{(3) Resource limitation}.  Despite the rapid hardware development, memory and computation power are still constrained on mobile devices and are orders of magnitude lower than their desktop and server counterparts.

Concluded from the challenges above, a good mobile inference engine should have the following two properties: (1) \emph{Universality} to address both model compatibility and device diversity; (2) \emph{Efficiency} to inference models on devices with great performance and to use as little memory and energy consumption as possible.

To satisfy above properties, we introduce a new mobile inference engine named \emph{Mobile Neural Network (MNN)}. Our contributions can be summarized as follows: 
\begin{itemize}
    \item We present a mechanism called pre-inference which manages to perform runtime optimization through online cost evaluation and optimal scheme selection.
    
    \item We deliver in-depth kernel optimization by utilizing improved algorithm and data layout to further boost the performance of some widely-used operations.

    \item We propose backend abstraction module to enable hybrid scheduling and keep engine itself as lightweight as possible. Integrating MNN into applications only increases the binary size by $400 \sim 600$KB.
\end{itemize}
    
It should be noted that MNN has been extensively adopted in many mobile applications. Meanwhile, we open-source the entire project to enrich the community and hope to involve more people to make it better.

\section{Related Work}
With the present rise of demand for on-device deep learning, much attention has been spent on mobile inference solutions, especially from several major companies.

CoreML is a framework of Apple to integrate machine learning models into iOS software applications, which can leverage multiple hardware like CPU, GPU, and ANE. For Android smartphones, Google also provides their own solution for on-device inference, \ie, ML-kit and Neural Networks API (NNAPI)~\cite{nnapi}. However, the main drawback of these solutions lies in their limited universality. For example, CoreML requires iOS 11+ and NNAPI requires Android 8.1+, which exclude many existing mobile phones and the embedded devices.

In late 2017, Google released TensorFlow Lite (TF-Lite)~\cite{tflite}, an efficient mobile deep learning framework. TF-Lite is optimized for less powerful devices such as mobile phones and embedded devices. Almost at the same time, Facebook released Caffe2~\cite{pytorch} to help developers deploy deep learning models on mobile devices. Both of them support a wide range of devices and many applications have already been developed based on them. 
However, on-device inference with TF-Lite or Caffe2 may sometimes go against the goal to be lightweight. For example, TF-Lite utilizes Accelate, Eigen, and OpenBLAS libraries for floating-point acceleration while Caffe2 depends on Eigen to accelerate matrix operations. Integrating mobile deep learning frameworks with these dependencies will bloat the binary size of mobile applications and bring in unnecessary overhead. 

Many efforts have been devoted to solving the problem above. NCNN~\cite{ncnn}, MACE~\cite{mace}, and Anakin~\cite{anakin} are the representatives. They follow a paradigm of what we call \emph{manually search} or \emph{non-automated search}. In this paradigm, operators are optimized \emph{case by case} through carefully-designed assembly instructions and do not rely on any external libraries. For example, a unique program function is implemented for a $3\times3$ convolution of stride $1$; another function has to be implemented separately for a $3\times3$ convolution of stride $2$. This kind of implementation allows the mobile inference engine to be lightweight and efficient. However, the case-by-case optimization is time-consuming and can hardly cover all new emerging operators.

In stark contrast to the manual search, there is another philosophy on the opposite, which we call \emph{automated search}, pioneered by TVM~\cite{tvm}. TVM is an open deep learning compiler stack that can compile various deep learning models into libraries in an end-to-end manner. Not only does it resolve the redundant dependency problem, but also provides graph-level and operator-level optimization customized for the model and backend. As a result, the performance of TVM is pretty encouraging and scalable in terms of both model and device diversity. However, these advantages come at a price. The runtime library generated by TVM is \emph{model-specific}, which means if we want to update the model (which can be very common and frequent for many AI-driven software applications), we need to re-generate the code and release a new version. This mechanism bears a burden of cost and sometimes it is impracticable for mobile applications. In this paper, we are motivated to develop a semi-automated search architecture featured by enhanced universality and better performance in mobile deployment.

\begin{figure*}[t]
    \centering
    \includegraphics[width=0.95\linewidth]{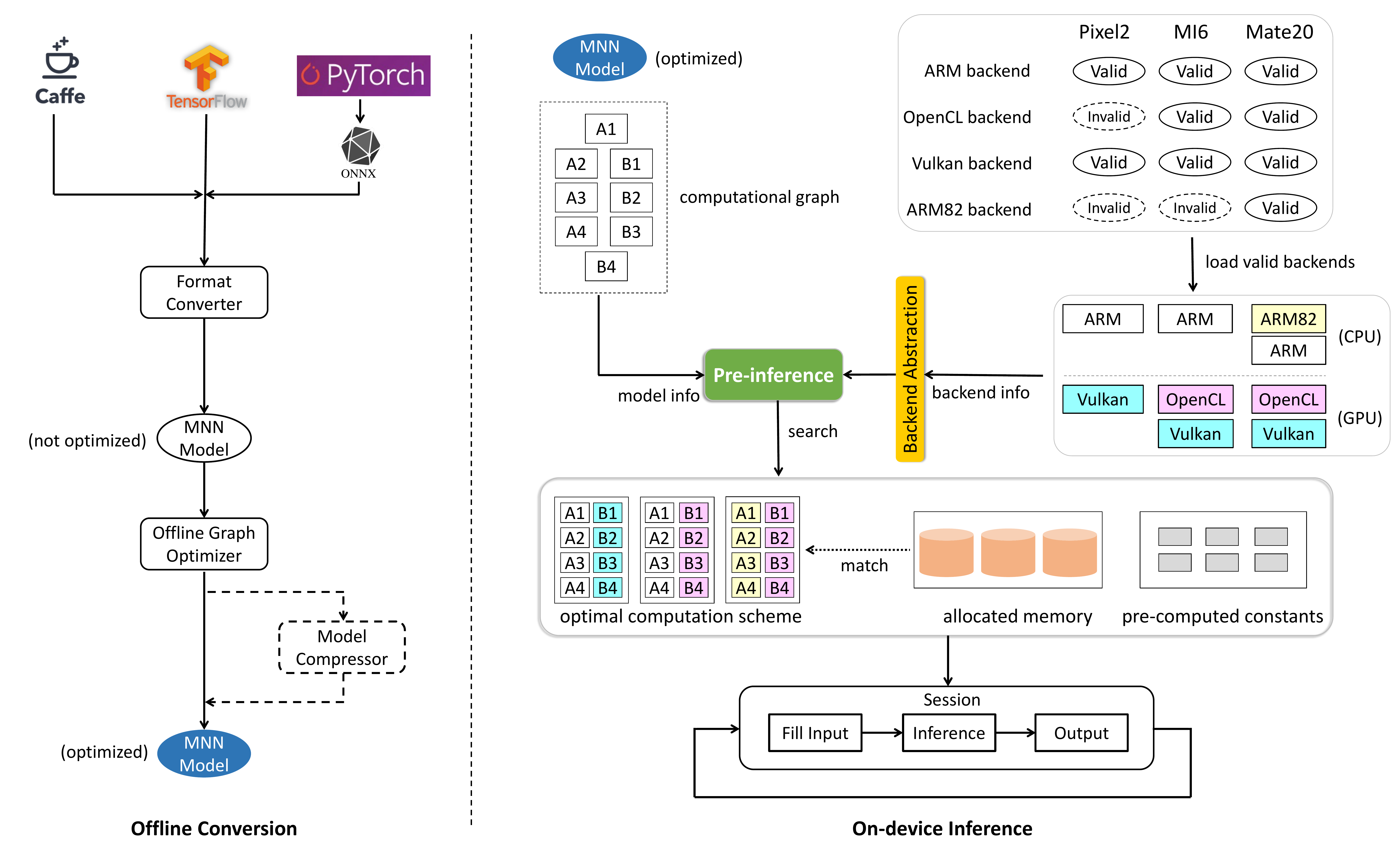}
    \caption{Architecture detail of the proposed mobile inference engine Mobile Neural Network (MNN).}
    \label{fig:mnn_architecture}
\end{figure*}

It is worthy of mention that there are some other works related to on-device deep learning, \eg, computational graph DSL (Domain-Specific Language)~\cite{abadi2016tensorflow,bastien2012theano} to perform optimization in the graph level or operator fusion and replacement~\cite{xla, wei2017dlvm}. These works are orthogonal to our contributions in this paper and are partially referenced by MNN.

\section{Mobile Neural Network (MNN)}
\subsection{Overview of MNN}

The basic workflow of MNN is composed of two parts, offline conversion and on-device inference. Figure~\ref{fig:mnn_architecture} summarizes the entire architecture of MNN and this section briefly shows how MNN optimizes the workflow by walking through its components. 

For the part of offline conversion, the converter first takes models as input from different deep learning frameworks and transforms them to our own model format (\verb+.mnn+). Meanwhile, some basic graph optimizations are performed, such as operator fusion~\cite{ashari2015optimizing}, replacement, and model quantization~\cite{rastegari2016xnor}.

For on-device inference, three modules are involved: pre-inference, operator-level optimization, and backend abstraction. For each operator, the pre-inference module offers a \emph{cost evaluation mechanism}, which combines the information (\eg, input size, kernel shape) with backend properties (\eg, number of cores, hardware availability) to dynamically determine the optimal computation solution from a solution pool. Then the operator-level optimization module utilizes advanced algorithms together with techniques like SIMD (Single Instruction Multiple Data), pipelining to further boost the performance.

Moreover, MNN supports various hardware architectures as backend. Since no single standard fits all hardware specifications, MNN supports different software solutions such as OpenCL, OpenGL, Vulkan, and Metal. All backends are implemented as independent components and a set of uniform interface is provided with backend abstraction module to hide raw details (\eg, memory management on heterogeneous backends).

With the proposed architecture, not only do we achieve high performance for on-device inference, but also make it easy to extend MNN to more ongoing backends (such as TPU, FPGA, \etc.). In the rest of this section, we present more details of the architecture of MNN.

\subsection{Pre-inference}
\label{subsec:pre-inference}
Pre-inference is the fundamental part of the proposed semi-automated search architecture. It takes advantage of a common phenomenon that the input size is typically fixed (or can be pre-processed into a target size) in many deep learning applications. Based on this, memory usage and computational cost can be determined \emph{ahead of} formal inferences. Thus, some optimization techniques like memory pre-allocation and reuse can be conducted to further improve the performance. Details of pre-inference can be specified into two parts: computation scheme selection and preparation-execution decoupling.

\textbf{Computation scheme selection}.
We propose a \emph{cost evaluation mechanism} to select the optimal scheme from the scheme pool, which takes into consideration both the algorithm implementation and the backend characteristics,
\begin{equation}
    C_{\text{total}} = C_{\text{algorithm}} + C_{\text{backend}},
\label{eqn:costevaluation}
\end{equation}
where $C$ stands for the cost.

(1) Take convolution scheme pool as an example, in which there are generally two fast implementation algorithms to choose: sliding window and Winograd~\cite{lavin2016fast}. Our general idea is to dynamically choose the algorithm that minimizes the computational cost based on different convolution settings. Therefore, the optimal computation scheme to minimize the $C_{\text{algorithm}}$ term can be determined as follows: 
\begin{enumerate}
    \item If kernel size $k = 1$, it is just a matrix multiplication. Strassen algorithm~\cite{strassen1969gaussian} will be applied.
    \item If kernel size $k > 1$, we employ Winograd to transform convolution into matrix multiplication. Winograd convolution has many sub-options for different output tile size $n$, with regard to which, the arithmetic cost can be formulated as follows. Based on the cost, we choose the optimal output tile size $\hat{n}$ that minimizes the cost $C$,
\begin{equation}
    \begin{aligned}
        C(n) &= 2 i_c (n+k-1)^3  \\
                &+ i_c o_c (n+k-1)^2 \\ 
                &+ n(n+k-1)(2n+k-1), \\
    \Rightarrow \hat{n} &= \arg \min_n C(n).
    \end{aligned}
\label{eqn:wino_cost}
\end{equation}

    \item Then, if $\hat{n}$ equals to $1$, sliding window is picked out; otherwise, we adopt Winograd convolution.
    
\end{enumerate}
This cost evaluation mechanism for convolution is summarized as
\begin{equation}
    \text{Scheme} = \left \{
    \begin{aligned}
       & \text{sliding window}, &\text{if} \; k > 1 \; \text{and} \; \hat{n} = 1, \\
       & F(\hat{n} \times \hat{n}, k\times k),  &\text{if} \; k > 1 \; \text{and} \; \hat{n} > 1. \\
    \end{aligned}
    \right.
\label{eqn:optimal_scheme}
\end{equation}

\begin{figure}[t]
    \centering
    \includegraphics[width=0.98\linewidth]{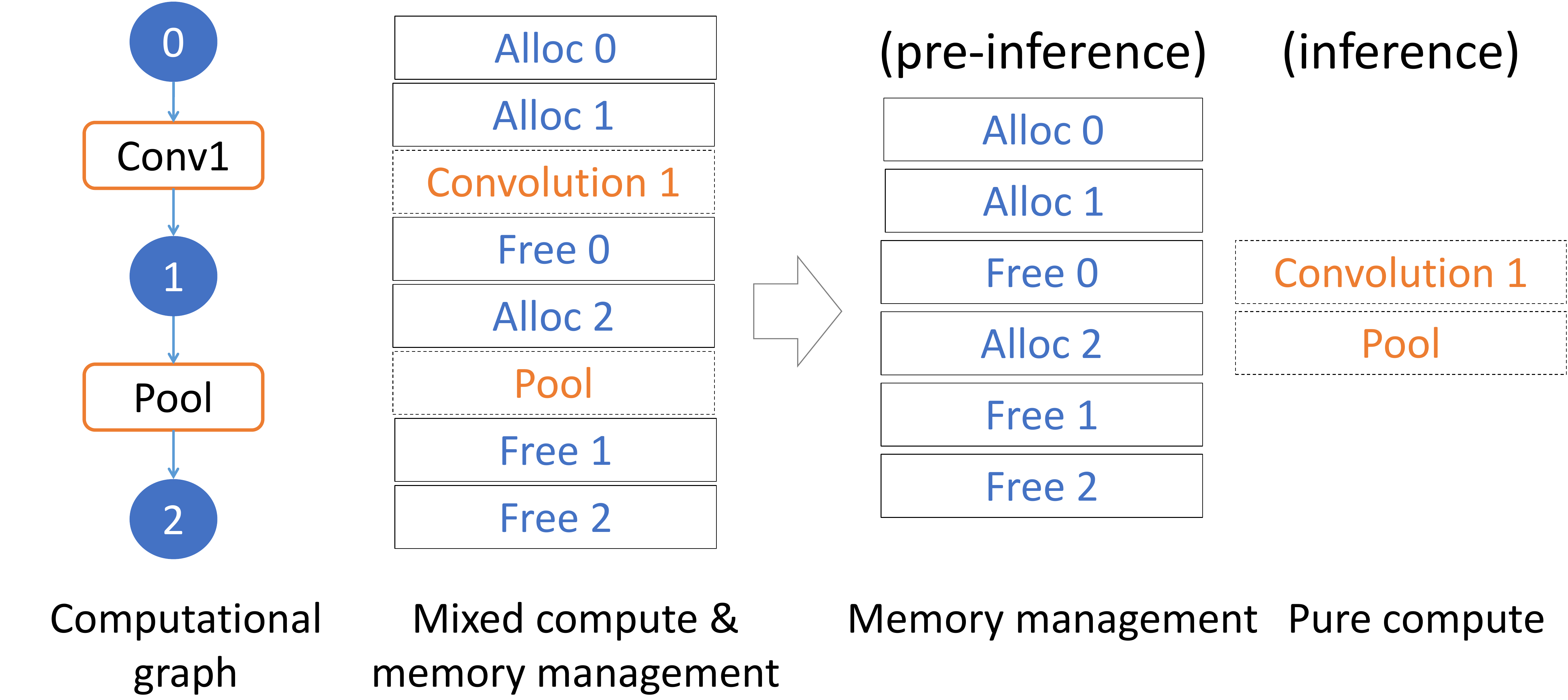}
    \caption{Memory optimization of MNN: decouple memory allocation from computation.}
    \label{fig:memory_allocation}
    \vspace{-1em}
\end{figure}

(2) Then next question is how we determine the second term $C_{\text{backend}}$ in Equation~\ref{eqn:costevaluation}. Generally, the idea is to sum up the time cost of all the operators on different backends, then choose the backend with the minimal cost,
\begin{equation}
    \begin{aligned}
        C_{\text{backend}} = \sum_{\text{op}} C_{\text{op}},
    \end{aligned}
\label{eqn:backend_cost}
\end{equation}
where $C$ denotes the cost as before and $\text{op}$ represents the operator. The key of backend cost evaluation is apparently to obtain the $C_{\text{op}}$ term, which relies on the different backend cases. If an operator is not supported on a GPU backend of interest (\eg, OpenCL/OpenGL/Vulkan), it will be scheduled to run on CPU. The cost of an operator on CPU or GPU can be formulated as
\begin{equation}
    C_{\text{op}} = \left \{
    \begin{aligned}
       & \frac{\texttt{MUL}}{\texttt{FLOPS}} \times 1000, & \text{if on CPU}, \\
       & \frac{\texttt{MUL}}{\texttt{FLOPS}} \times 1000 + t_{\text{schedule}}, & \text{if on GPU}, \\
    \end{aligned}
    \right.
\label{eqn:op_cost}
\end{equation}
where $\texttt{MUL}$ stands for the number of multiplication, indicating the computational complexity of an operator. \texttt{FLOPS} (FLoating-Operations Per Second) is a common index to quantify the computation power of CPU or GPU. As we see, the main difference of GPU from CPU as the backend is that it has an additional term $t_{\text{schedule}}$, which accounts for the cost to prepare the command buffer and command description for GPU. Note that, \texttt{FLOPS} and $t_{\text{schedule}}$ are known constants for a specific backend (please refer to the Appendix for how to determine them in detail). 

Table~\ref{tab:my_label} shows the performance comparison between fixed scheme and our scheme selection under different convolution settings. We can see that each scheme of sliding window or Winograd can be especially fit in certain case, but not universally good. Our method gets the best (or comparable to the best) in different cases, partly showing the claimed universal efficiency.

\begin{table}[t]
    \centering
    \caption{Inference time (ms) comparison of different computation schemes. "Sliding" represents "sliding window" and "WinoMin/Max" means "Winograd convolution with minimal/maximal block size". The four numbers in the convolution argument setting successively mean kernel size, input channel, output channel, and the spatial size of input, respectively. The best results are \bb{in bold} and the second best are \U{underlined}.}
    \vspace{0.2em}
    \setlength\tabcolsep{2.8pt} 
    \begin{tabular}{|c|c|c|c|}
    \hline 
    \multirow{2}*{Scheme} & \multicolumn{3}{c|}{Convolution argument setting} \\
    \cline{2-4}
    & \small $(2,3,16,224)$ & \small $(2,512,512,16)$ & \small $(3,64,64,112)$ \\
    \hline \hline
    Sliding     & $\mathbf{32.1}$ & $895.1$ & $895.1$ \\ 
    WinoMin     & $42.2$ & $\U{287.7}$ & $389.8$ \\
    WinoMax     & $57.3$ & $539.3$     & $\U{237.4}$ \\
    Ours & $\U{32.7}$ & $\mathbf{286.0}$ & $\mathbf{236.4}$ \\
    \hline
    \end{tabular}
    \label{tab:my_label}
    \vspace{-1em}
\end{table}

\noindent \textbf{Preparation-execution decoupling}.
During the program execution, the computation is normally interlaced with the memory allocation and freeing. Speaking of mobile applications, however, time spent on memory management cannot be ignored. Since input size is determined or can be pre-processed into a target size, MNN can infer the exact required memory for the entire graph by virtually walking through all operations and summing up all allocation and freeing. In this sense, MNN pre-allocates required memory as a memory pool during the pre-inference stage and reuses it in the following inference sessions. The whole process is illustrated in Figure~\ref{fig:memory_allocation}.

It should also be noted that by decoupling preparation from execution, MNN achieves better performance when the selected scheme is GPU-related. This is because setting up command buffer and its related command descriptions is, to some extent, time-consuming and thus have negative impact on the inference performance.

This simple idea can be quite effective in practice. The inference time using this technique can drop by about $7\% \sim 8\%$ on CPU and $50\% \sim 75\%$ on GPU. More details can be found in Table~\ref{tab:result_decoupling}.

\begin{table}[t]
    \centering
    \caption{Inference time (ms) comparison without (w/o) and with (w/) the proposed preparation-execution decoupling. Hardware setting: (1) MI6 -- CPU: Kryo 280, GPU: Adreno 540; (2) P10 -- CPU: Cortex A73, GPU: Mali G71.}
    \vspace{0.2em}
    \begin{tabular}{|c|c|c|}
        \hline
        Scheme & CPU ($4$ threads) & GPU (Vulkan) \\
        \hline \hline
        MI6 (w/o)   & $30.9$                      & $63.6$ \\
        MI6 (w/)    & $\mathbf{28.9}$ ($\downarrow6.5\%$)  & $\mathbf{15.8}$ ($\downarrow75.2\%$) \\
        \hline \hline
        P10 (w/o)   & $29.0$                      & $41.0$ \\
        P10 (w/)    & $\mathbf{26.8}$ ($\downarrow7.6\%$)  & $\mathbf{20.7}$ ($\downarrow49.5\%$) \\
        \hline
    \end{tabular}
    \label{tab:result_decoupling}
    \vspace{-1em}
\end{table}

\subsection{Kernel Optimization}
\label{subsec:kernel_optim}
Kernel is the detailed implementation of an operator, whose optimization can be specified into two ways, \ie, the algorithm and the schedule~\cite{jiang2018efficient}. In other words, we need to choose the optimal algorithm with the lowest arithmetic complexity and make the most of the available hardware resources for fastest execution.

\subsubsection{Winograd optimization}
\label{subsubsec:winograd}
\noindent Winograd algorithm is a well-known minimal filtering algorithm proposed by Shmuel Winograd~\cite{winograd1980arithmetic} and has been used to accelerate convolution in DNNs~\cite{lavin2016fast}. 

Given an input feature map of size $[i_w, i_h, i_c]$, its output feature map $[o_w, o_h, o_c]$, Winograd convolution can be formulated as 
\begin{equation}
    Y = A^T \left[ \sum_{\text{channel}}(GWG^T) \odot (B^TXB) \right] A,
\label{eqn:winograd}
\end{equation}
where $G, B, A$ are the transformation matrices for kernel $W$ (spatial size $[k, k]$), input $X$ (spatial size $[n+k-1, n+k-1]$), and output $Y$ (spatial size $[n, n]$), respectively. These three matrices only depend on the shape of $W$ and $X$.

We optimize Winograd convolution based on a proposed Winograd generator with popular parallel computing techniques such as pipelining and SIMD.

\noindent \textbf{(1) Block division and pipelining}.
In Winograd convolution~\cite{lavin2016fast}, $X$ is a small tile instead of the whole input feature map, thus leaving us the first issue: block division, \ie, how to determine $n$.

To resolve it, we divide the block from the \emph{output} perspective. For the output of size $[o_w, o_h, o_c]$, let $T$ be the multiplier for parallel computing (namely, we compute $T$ output blocks per time). Then there should be
\begin{equation}
    T =  \left \lfloor \frac{o_w o_h}{\hat{n}^2} \right \rfloor,
    \label{eqn:computation_scheme}
\end{equation}
where $\hat{n}$ is the aforementioned optimal output tile size decided at the pre-inference stage (Equation~\ref{eqn:wino_cost}).

When blocks are computed together, we must try our best to avoid pipeline stalls to hide latency. The trick is well-known as avoiding data dependence in pipeline~\cite{kennedy2001optimizing}. This is achieved by careful \emph{assembly instruction rearrangement} in MNN.

\noindent \textbf{(2) Hadamard product optimization}. Hadamard product is an essential step in Winograd convolution (see Equation~\ref{eqn:winograd}). However, it has a problem that memory access takes much time, draging down the whole acceleration. 

\begin{figure}[t]
    \centering
    \includegraphics[width=\linewidth]{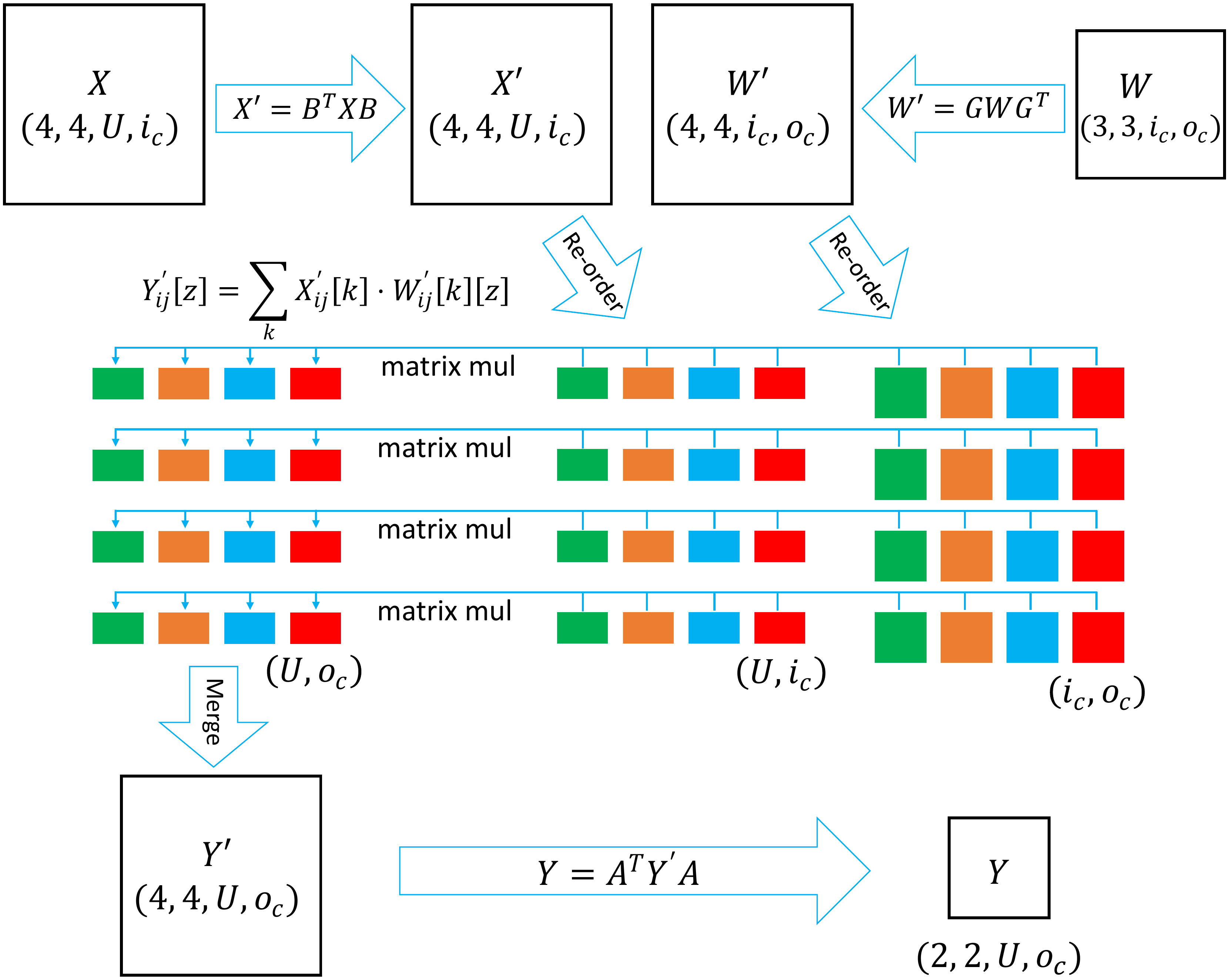}
    \vspace{-1.5em}
    \caption{Illustration of the optimized Winograd algorithm in MNN (\emph{best viewed in color}).}
    \label{fig:winograd}
\end{figure}

From Equation~\ref{eqn:winograd}, it can be found that the summation plus the Hadamard product can be transformed into a dot product. Combining many dot product together gives us matrix multiplication, which is a good indicator for parallelism and amortizing memory access overhead. In this spirit, we propose to transform the Hadamard product to matrix multiplication building upon \emph{data layout re-ordering}. The consequent new data layout is called NC4HW4~\cite{tvm,metal}. Briefly, NC4HW4 is a data layout re-ordering method that splits out $V$ ($V=4$ in this paper) data elements as a unit to make a new dimension for a tensor. The $V$ elements are placed contiguously in memory so as to leverage the vector registers in CPUs to compute $V$ data in a single instruction (\ie, SIMD). After this re-ordering, the Winograd convolution is illustrated as Figure~\ref{fig:winograd}.

\noindent \textbf{(3) Winograd generator}. Most existing inference frameworks using Winograd~\cite{tflite,ncnn,mace} hardcode the $A, B, G$ matrices for common kernel and input sizes in their source codes~\cite{tfwinograd,macewinograd}, which has relatively poor scalability in face of new cases. In contrast, MNN keeps a universal interface through a proposed \emph{Winograd generator}, allowing for Winograd convolution of \emph{any} kernel and input sizes. We adopt the following formula to generate the $A, B$ matrices,
\begin{equation}
  \begin{aligned}
    x \cdot (x-f)(x+f) \cdot (x-2f)(x+2f) \cdots \\
    (x-\frac{(n+k-1)f}{2})(x + \frac{(n+k-1)f}{2}),
  \end{aligned}
\label{eqn:wino_generator}
\end{equation}
where $f$ is a scalar used to minimize the numerical errors. We set $f=0.5$ in this paper. 

\subsubsection{Large matrix multiplication optimization}
As aforementioned (Section~\ref{subsec:pre-inference}), convolution operations with kernel size $1$ are converted into large matrix multiplication in MNN, where comes the Strassen algorithm~\cite{strassen1969gaussian} for acceleration. To the best of our knowledge, MNN is the \emph{first} mobile inference engine to adopt Strassen algorithm to accelerate large matrix multiplication. 

Strassen is a fast algorithm that trades expensive multiplications with cheap additions, whose acceleration effect is maximized when it is applied \emph{recursively}~\cite{blahut2010fast}. In practice, we need to decide when the recursion should stop. Conditioned on a fact that on modern processors, multiplication is nearly the same costly as addition, so we can just compare their cost via their numbers. In MNN, for a matrix multiplication of size $[n, k] \times [k, m] \Rightarrow [n, m]$, the number of direct multiplication is $mnk$, while it only needs $7 \cdot \frac{m}{2}\frac{n}{2}\frac{k}{2}$ multiplications using Strassen. The extra cost of using Strassen is $4$ matrix additions of size $[\frac{m}{2}, \frac{k}{2}]$, $4$ matrix additions of size $[\frac{n}{2}, \frac{k}{2}]$, and $7$ matrix additions of size $[\frac{m}{2}, \frac{n}{2}]$. Therefore, the recursion goes on only when the benefit is over cost, formulated as
\begin{equation}
    mnk - 7 \cdot \frac{m}{2}\frac{n}{2}\frac{k}{2} > 4 \cdot \frac{m}{2}\frac{k}{2} + 4 \cdot \frac{n}{2}\frac{k}{2} + 7 \cdot \frac{m}{2}\frac{n}{2}.
\label{eqn:strassen_cost}
\end{equation}
Once this inequation cannot hold, the recursion of Strassen should stop.

Table~\ref{tab:result_Strassen} demonstrates the advantage of Strassen compared with direct matrix multiplication with different matrix sizes. We can see that the Strassen method outperforms the direct one by $7.5\% \sim 13.5\%$ improvement.

\begin{table}
    \centering
    \caption{Time cost (ms) of matrix multiplication comparison on P10 without (w/o) and with (w/) the optimized Strassen algorithm. Matrix size of (a, b, c) means the matrix multiplication of size [a, b] times [b, c].}
    \vspace{0.2em}
    \begin{tabular}{|c|c|c|}
        \hline
        Matrix size & w/o Strassen & w/ Strassen \\
        \hline \hline 
        $(256, 256, 256)$ & $23$ & $23$ \\
        $(512, 512, 512)$ & $191$ & $\mathbf{176}$ ($\downarrow7.9\%$) \\
        $(512, 512, 1024)$ & $388$ & $\mathbf{359}$ ($\downarrow7.5\%$) \\
        $(1024, 1024, 1024)$ & $1501$ & $\mathbf{1299}$ ($\downarrow13.5\%$) \\
        \hline
    \end{tabular}
    \label{tab:result_Strassen}
\end{table}

\subsection{Backend Abstraction}
\label{subsec:backend}
Backend abstraction module is introduced to make all the hardware platforms (\eg, GPU, CPU, TPU) and software solutions (\eg, OpenCL, OpenGL, Vulkan) encapsulated into a uniform \verb+Backend+ class. Through \verb+Backend+ class, resource management, memory allocation, and scheduling are disentangled with the concrete operator implementations.

\begin{figure}[t]
    \centering
    \includegraphics[width=\linewidth]{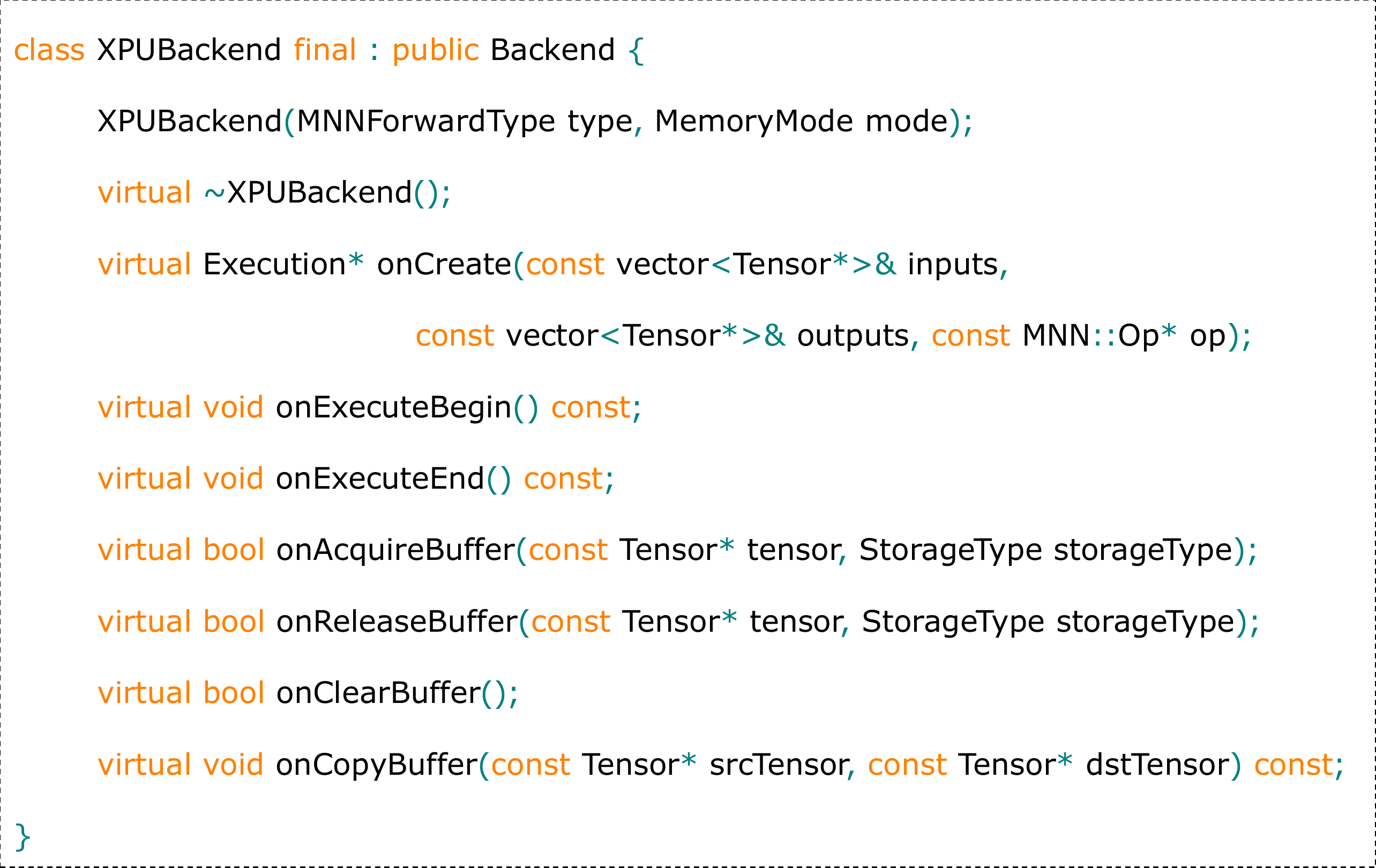}
    \vspace{-1.5em}
    \caption{Backend class in MNN (\emph{best viewed in color}).}
    \label{fig:backend_code}
\end{figure}

The \verb+Backend+ class consists of several abstract functions, as shown in Figure~\ref{fig:backend_code}. As for memory management, \verb+onAcquireBuffer+ is responsible for allocating new memory for tensors and \verb+onReleaseBuffer+ for releasing them. For operator implementation, \verb+onCreate+ is designed to create execution instance for each operator. 

Advantages of this specific module are three-folds.

\begin{figure*}[t]
    \centering
    \begin{tabular}{ccc}
      \includegraphics[height=0.24\linewidth]{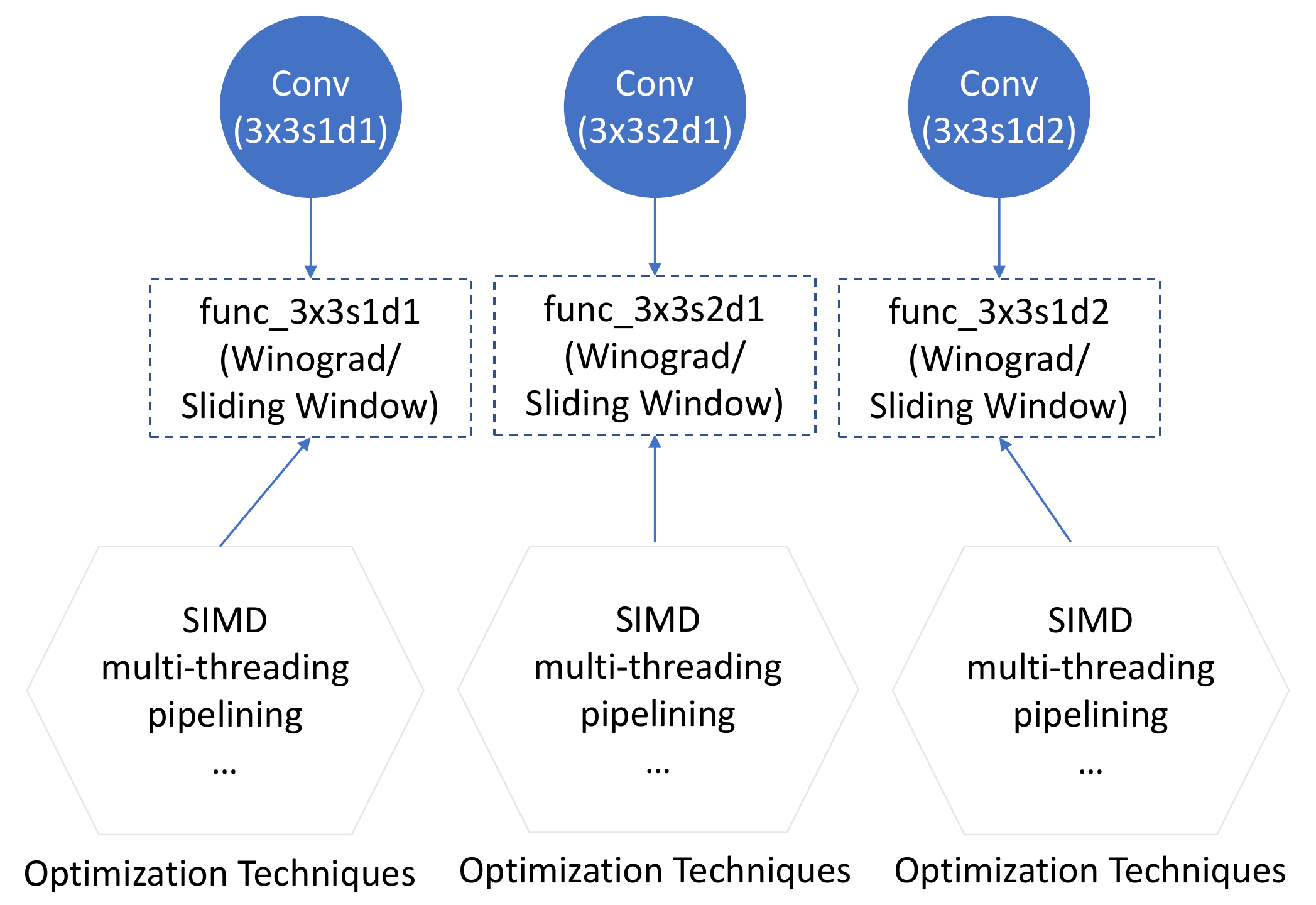} &
      \includegraphics[height=0.24\linewidth]{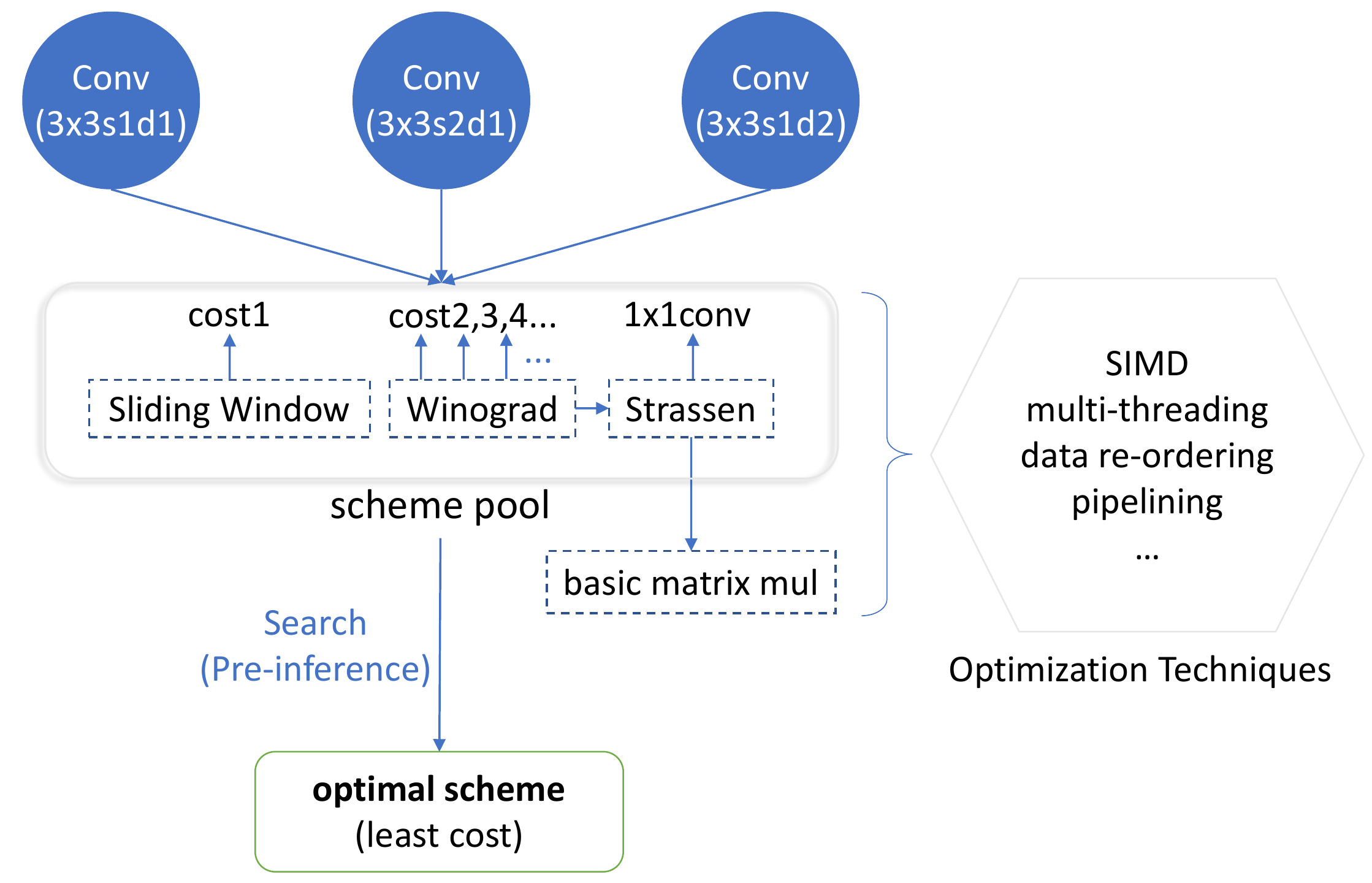} &
      \includegraphics[height=0.24\linewidth]{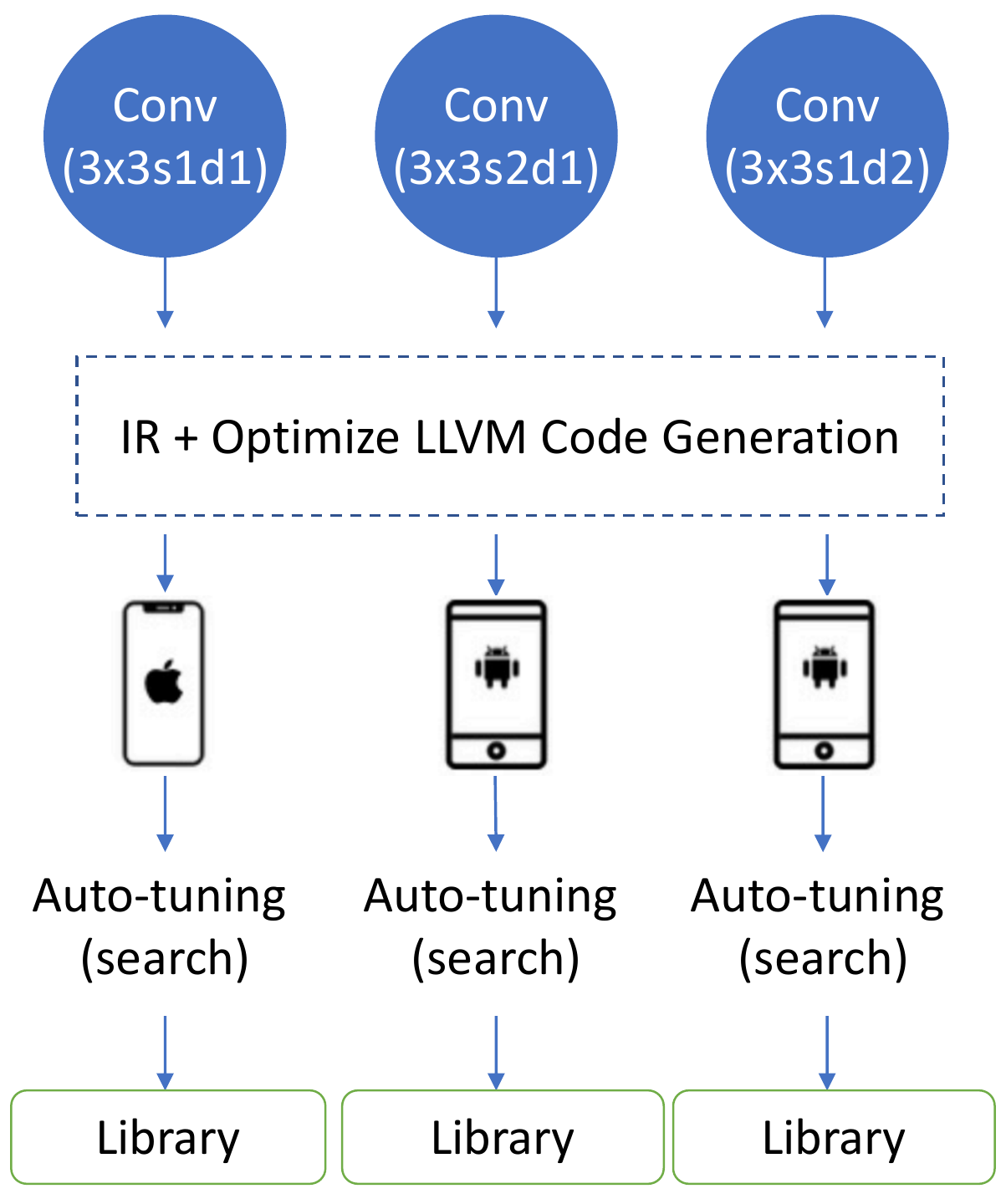} \\
      \vspace{0.05em} \\
      \small (a) Manual Search & \small (b) Semi-automated Search (MNN)  & \small (c) Automated Search \\
    \end{tabular}
    \caption{Design paradigm comparison between MNN and the other two kinds of popular mobile inference engines.}
    \label{fig:paradigm_comparison}
    \vspace{-1em}
\end{figure*}

\textbf{(1) Reduce complexity}. A large number of operators and innumerable devices make the operator optimization a non-trivial task. The major challenge lies in the fact that heterogeneous backends typically have different ways to manage resources, \eg, allocating/freeing memory and scheduling data. Dealing with these issues makes implementation error-prone and cumbersome. The \verb+Backend+ class uniformly manages resource loading such as GPU shader and completes the optimal memory allocation according to the declarations of operators. With backend abstraction, MNN manages to divide the task into two independent parts. The "front-end operator" developers can focus on the efficient implementation for fast operator execution and all unrelated backend details are hidden. The "backend" developers can be devoted to exploiting different backend specifications and offering more convenient APIs. This separation of tasks are rather meaningful in practice since the lowering contribution barriers is highly appreciated in open-source projects.

\textbf{(2) Enable hybrid scheduling}. Heterogeneous computing mainly involves the backend selection and data transmission among different backends. When creating an inference session in MNN, one can configure targeted backends. If there are multiple backends available, MNN can decide the optimal backends for operators according to aforementioned backend evaluation (Section~\ref{subsec:pre-inference}). As a result, MNN supports the flexible combination of operator execution on different backends even in a single inference. For example, convolution may run on CPU and the following ReLU activation may run on GPU. With the help of \verb+Backend+ class, developers do not need to worry about the scheduling and transmission, which automatically proceed under the hood.

\textbf{(3) More lightweight}. Implementation of each backend can work as an independent component while maintaining the uniform interface in MNN. Whenever some backends are unavailable on certain device, the corresponding  concrete implementation can be easily peeled off from the whole framework. For example, Metal~\cite{metal} is not supported on Android platforms and thus we can just take away the Metal module without touching the specific operator implementations. Through the decoupling of operator and backend, MNN can be competitively lightweight. This is of critical importance since mobile applications have a rigorous restriction on the binary size.

Through these uniform backend interfaces, to the best of our knowledge, MNN is the inference engine that supports the \emph{most comprehensive} backends (see Table~\ref{tab:engine_comparison}). Moreover, it is scalable enough for users to integrate new backends such as NPU, FPGA, \etc.

\subsection{Method Summary}
\label{subsec:big_picture_compare}
As compared with some other typical design paradigms shown in Figure~\ref{fig:paradigm_comparison}, we illustrate the design philosophy behind MNN and highlight several MNN-specific advantages in this part.

For an inference engine, high performance is the vital factor to decide whether it will be adopted by developers or not. Out of this motivation, different inference engines (\eg, TF-Lite, NCNN, TVM)  put continuous efforts into optimization to achieve better performance in difference ways.

For MNN, we make many improvements to meet the fundamental requirement of high performance. Taking the overwhelming operator diversity into consideration, we are not satisfied with the case-by-case optimization solution. Although this solution can be rather simple and effective, it often encounters the problem that some operators are left out of optimization and become the performance bottleneck (as shown by an example in Section~\ref{subsec:experimental_results}). Instead, MNN first spots the compute-intensive unit \emph{of a smaller granularity} (\ie, the basic matrix multiplication), which is highly optimized by means of fast algorithms and paralleling techniques. Consequently, operators built upon this basic unit can naturally benefit from acceleration without being specially optimized.

\begin{table*}[t]
    \centering
    \begin{threeparttable}[b]
    \caption{Backend comparison of different mobile inference engines. "--" means that the mobile engine does not support that kind of backend (at the time this paper is submitted); "/" stands for "not applicable". Metal~\cite{metal} is Apple's exclusive GPU standard on iOS, while OpenCL~\cite{opencl}, OpenGL~\cite{opengl}, and Vulkan~\cite{vulkan} are the concurrent standards on Android.}
    \vspace{0.5em}
    \begin{tabular}{|c|c|c|c|c|c|c|}
    \hline
    \multirow{2}*{Mobile Inference Engine} & \multirow{2}*{\#Operator (CPU)}  & \multicolumn{4}{c|}{\#Operator (GPU)} & \multirow{2}*{Supported OS} \\
                                                         \cline{3-6}
                                &                           & Metal & OpenGL & OpenCL & Vulkan & \\ 
    \hline \hline
    CoreML~\cite{coreml}        & $110$\tnote{1}             & $110$\tnote{1} & / & /      & /      & iOS \\
    TF-Lite~\cite{tflite}       & $93$                       & $17$  & $19$   & --     & --     & iOS+Android \\
    MACE~\cite{mace}            & $61$                       & --    & --     & $29$   & --     & Android \\
    NCNN~\cite{ncnn}            & $65$                       & --    & --     & --     & $32$   & iOS+Android \\
    MNN (Ours)                  & $94$                       & $55$  & $15$   & $33$   & $35$   & iOS+Android \\
    \hline
    \end{tabular}
    \label{tab:engine_comparison}  
    \begin{tablenotes}
        \item[1] \small Since CoreML is not open-sourced, we cannot tell which operator belongs to CPU or GPU. Thus the number here is the total number of operators.
    \end{tablenotes}
    \end{threeparttable}
    \vspace{-1em}
\end{table*}

Besides, maintainability, scalability, and the deployment cost all have a great impact on the long-term growth of an inference engine. Compared with auto-tuning in TVM, MNN is able to select the optimal computation scheme with less time and realize the runtime optimization through the proposed pre-inference mechanism. Note that, by transferring the search stage from offline compilation to online pre-inference, we also avoid the restriction on the binary validation (\eg, the iOS code signature). Meanwhile, by providing a set of uniform interface to hide raw backend details, MNN is naturally equipped with the property of modularity. Not only does this property make MNN lightweight, but also it is more convenient for contributors to extend MNN to more backends, as partly shown by number of operators supported on different backends in Table~\ref{tab:engine_comparison}.

\section{Benchmark Experiments}
\label{sec:exp}
In this section, we comprehensively evaluate the performance of MNN. We first explain our experiment settings, then present the experimental results on different hardware platforms and networks, compared with other mobile inference engines. Finally, we share an online case to show the production application of MNN.

\subsection{Experiment Settings}
\begin{itemize}
    \item Inference engines. We compare the performance with state-of-the-art mobile inference engines, including CoreML~\cite{coreml}, TF-Lite~\cite{tflite}, NCNN~\cite{ncnn}, and MACE~\cite{mace}.
    \item Devices. For iOS, we adopt iPhone8 and iPhoneX (processor: Apple A11 Bionic), as they are popularly adopted in benchmarks\footnote{https://www.tensorflow.org/lite/performance/benchmarks}. For Android, MI6 (processor: Snapdragon 835) and Mate20 (processor: Kirin 980) are adopted. 
    \item CPU and GPU. (1) For CPU, thread $\{2, 4\}$ are evaluated considering that modern devices usually have two or four processors and multi-threading is a common acceleration technique. CPU affinity is set to use \emph{all} the available cores in line with NCNN benchmark~\cite{ncnn}. (2) For GPU, we evaluate the Metal Performance Shaders on iPhones. On Android devices, three standard backends (\ie, OpenCL, OpenGL, and Vulkan) are evaluated since MNN has supported them all (see Table~\ref{tab:engine_comparison}).
    \item Networks. MobileNet-v1~\cite{howard2017mobilenets}, SqueezeNet-v1.1~\cite{IanMosAsh16}, and ResNet-18~\cite{HeZhaRenSun16} are chosen as benchmark networks since they have been extensively used in mobile applications.
    \item Run settings. We report the inference time of one $224\times224$ RGB image (\ie, batch size is $1$), averaged by $10$ runs. Before benchmark, one warm-up inference is conducted for fair comparison with other works~\cite{ncnn,tflite}.
\end{itemize}

\subsection{Experimental Results}
\label{subsec:experimental_results}
\begin{figure*}[t]
    \centering
    \begin{tabular}{c}
        \includegraphics[width=\linewidth]{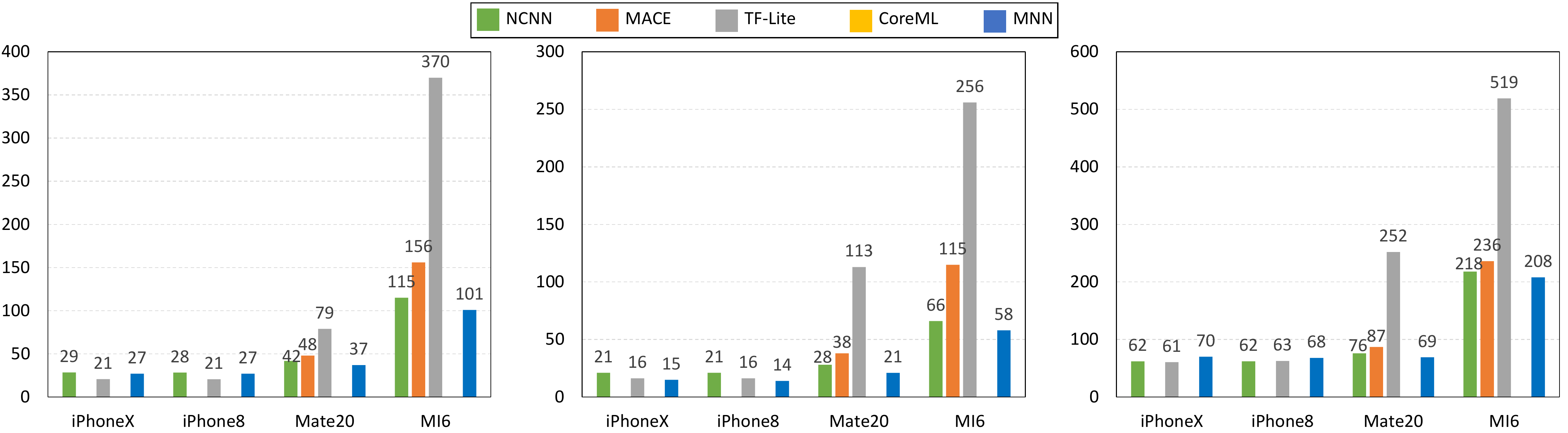} \\
        \includegraphics[width=\linewidth]{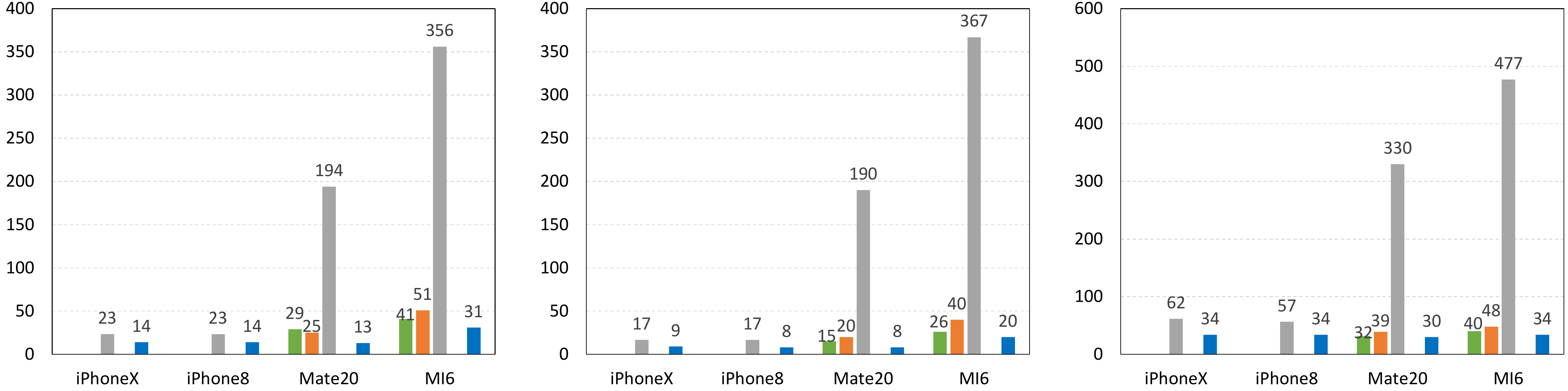} \\
        \includegraphics[width=\linewidth]{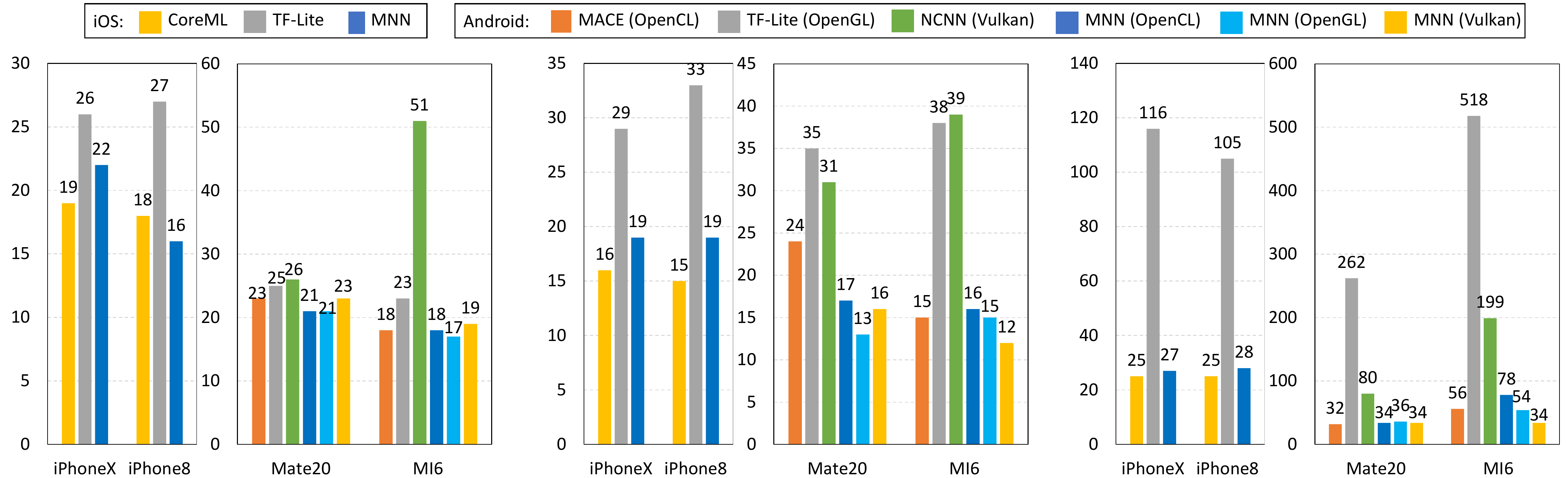} \\
    \end{tabular}
    \caption{Inference time (ms) comparison on MobileNet-v1 (left), SqueezeNet-v1.1 (middle), and ResNet-18 (right). Row 1: CPU with $2$ threads. Row 2: CPU with $4$ threads. Row 3: GPU. (\emph{Best viewed in color})}
    \label{fig:cpu_result}
    \vspace{-1.5em}
\end{figure*}

\textbf{Performance on different smartphones and networks}. Figure~\ref{fig:cpu_result} shows the performance of MNN compared with other four inference engines. We have the following observations.

(1) Generally, MNN outperforms other inference engines under \emph{almost all} settings by about $20\% \sim 40\%$, regardless of the smartphones, backends, and networks. 

(2) For CPU, on average, 4-thread inference with MNN is about $30\%$ faster than others on iOS platforms, and about $34\%$ faster on Android platforms (\eg, Mate20).

(3) For Metal GPU backend on iPhones, MNN is much faster than TF-Lite, a little slower than CoreML but still comparable, which is reasonable since CoreML is Apple's exclusive GPU solution tailored to iOS while MNN is meant to support backends of different operating systems. For Android GPU backends, other engines usually have their performance blind spots. For example, NCNN with Vulkan backend is not very fast on MI6; TF-Lite with OpenGL still has much room for improvement on ResNet-18 network. In contrast, MNN obtains favorable results on \emph{all} different hardware platforms and networks. Note that we achieve this comprehensive performance advantage by means of the proposed semi-automated search architecture rather than case-by-case heavy optimization.

(4) The multi-thread CPU inference using MNN on high-end devices (\eg, iPhone8 and iPhoneX) is highly competitive compared with that using GPU backends, which demonstrates the effectiveness of the proposed in-depth kernel optimization of MNN.

\begin{figure}[t]
    \centering
    \includegraphics[width=\linewidth]{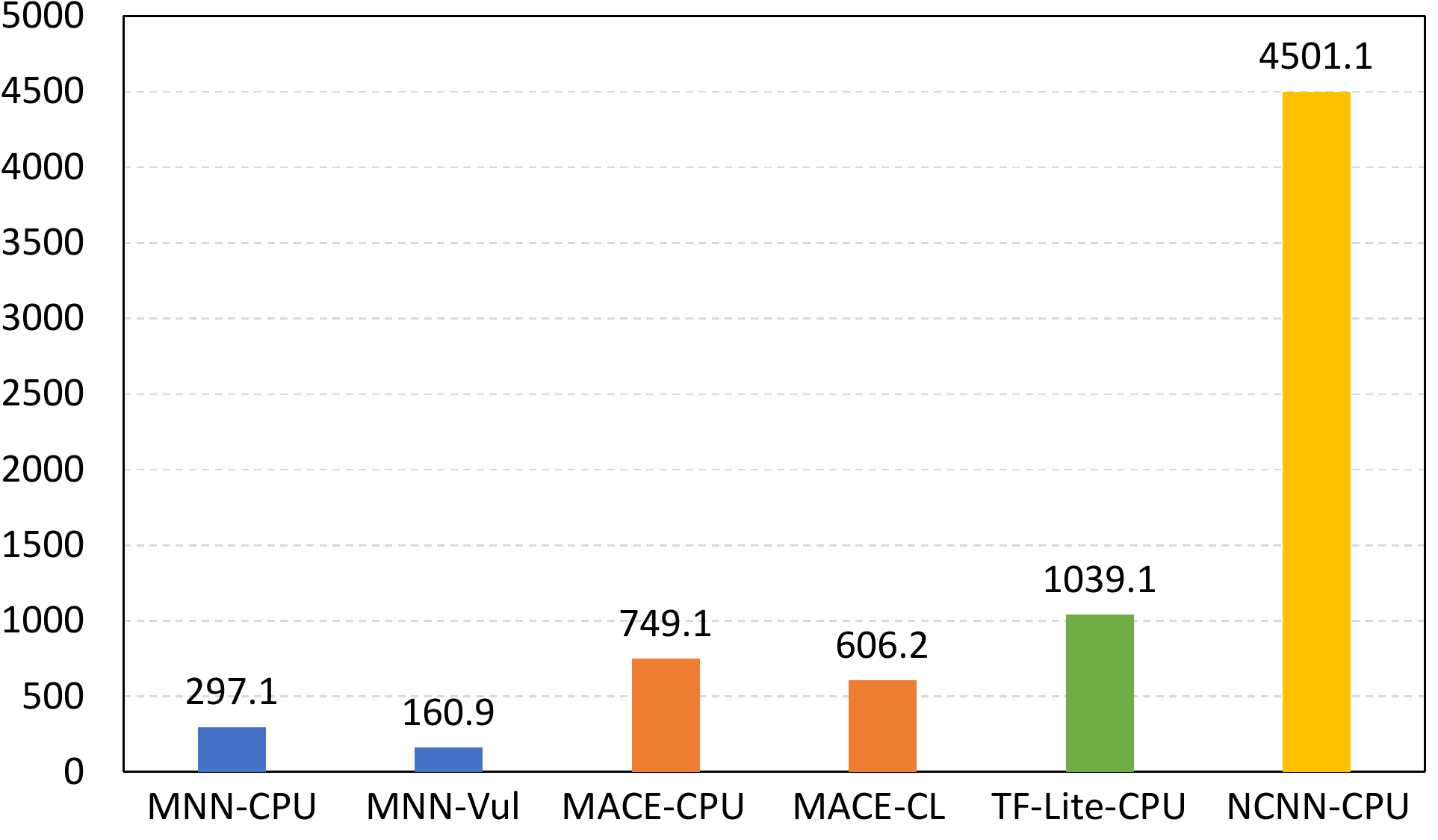}
    \vspace{-1.5em}
    \caption{Bottleneck of case-by-case optimization on Inception-v3, evaluated on Huawei P20 (Kirin 970). "MNN-Vul" means"MNN-Vulkan" and "MACE-CL" stands for "MACE-OpenCL".}
    \label{fig:bad_ncnn}
    \vspace{-1.5em}
\end{figure}

\textbf{Bottleneck of case-by-case optimization}. Figure~\ref{fig:bad_ncnn} shows an under-performance example of the case-by-case optimization on Inception-v3~\cite{SzeLiuJiaSerReeAngErhVanRab15}. One can clearly see that NCNN requires abnormally more inference time than the others. This is because some special operators in this network (\eg, $1\times7$ and $7\times1$ convolution) are not optimized by NCNN (for now). Consequently, they become bottlenecks during execution and drag down the overall performance severely. This specific case shows the limited scalability of case-by-case optimization. MNN is free from this problem because our computation solution is a general one which is applicable for various convolution cases.

\begin{figure}[t]
    \centering
    \includegraphics[width=\linewidth]{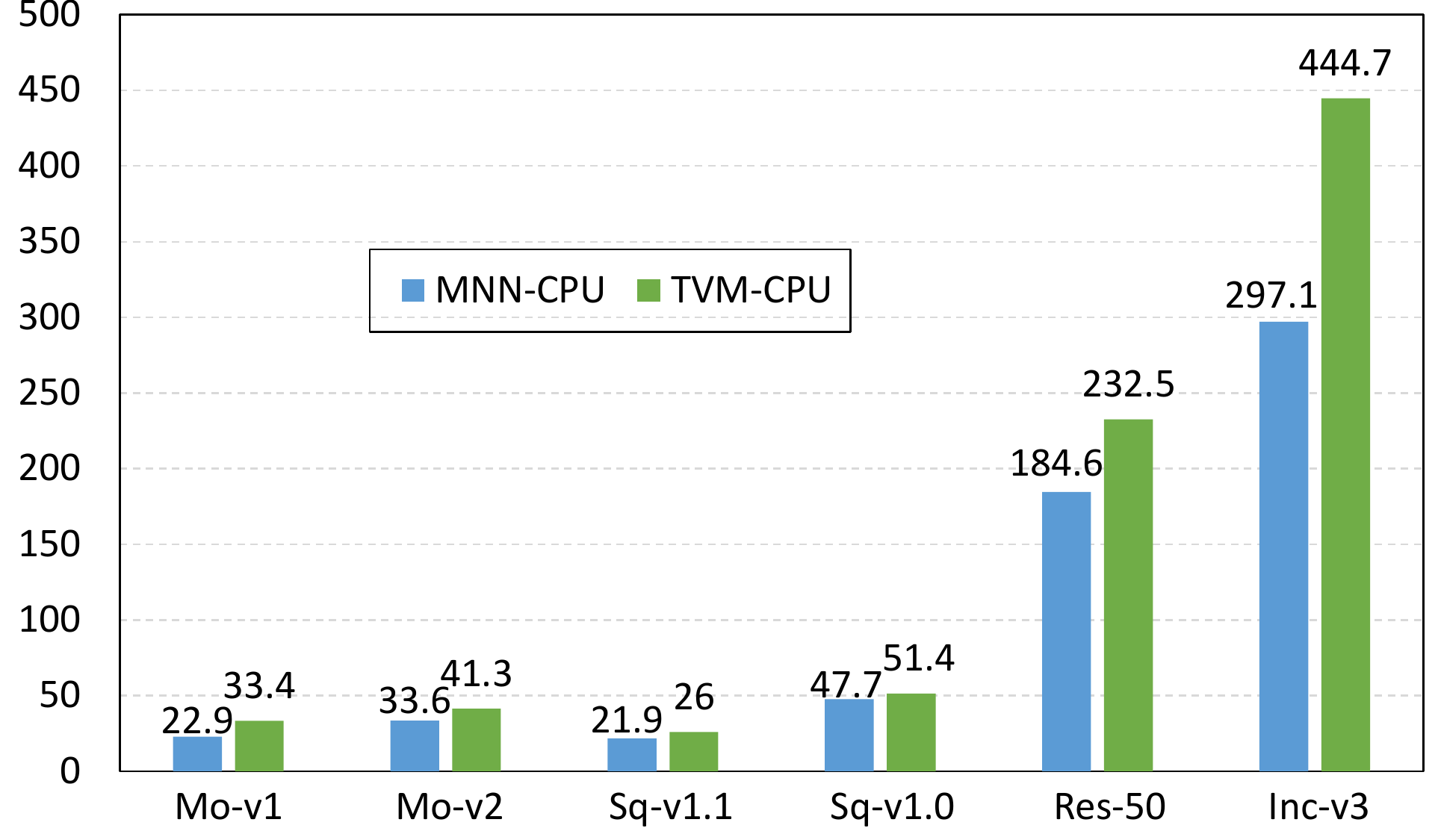}
    \vspace{-1.5em}
    \caption{CPU inference time (ms) comparison between MNN and TVM~\cite{tvm} on Huawei P20 Pro (SoC: HiSilicon Kirin 970). "Mo/Sq/Res/Inc" is short for MobileNet/SqueezeNet/ResNet/Inception, respectively. TVM data is adopted from their released benchmark: https://github.com/dmlc/tvm/wiki/Benchmark.}
    \label{fig:tvm_comparison}
    \vspace{-1.5em}
\end{figure}

\textbf{Comparison with TVM}. We compare the performance of MNN with TVM on various networks, as shown in Figure~\ref{fig:tvm_comparison}. Although MNN does not apply model-specific tuning and optimization, MNN still has encouraging performance and is even slightly faster than TVM. In addition, generating model-specific code using auto-tuning and compiling in TVM usually brings some deployment overhead. In Table~\ref{tab:tvm_time_cost}\footnote{One may feel curious why the compiling results are faster than auto-tuning. This is because, TVM provides developers with the mechanism (\ie, tensorize) to replace the auto-tuned code with hand-optimized implementation. Thus, it is reasonable for our optimized direct compilation to outperform auto-tuning.} we show the time cost of auto-tuning and compiling for ResNet-18 with TVM. Even with a small number of trials for tuning on a single device, TVM still takes much time to generate code. Since most mobile applications cover lots of different device types, the process of code generation will take much longer time and demand more resources (\eg, servers), which is hardly affordable for many developers. MNN is free from these issues because all optimizations are performed at runtime without performance loss.

\begin{table}[t]
    \centering
    \caption{Time cost (s) of auto-tuning and compiling for ResNet-18 with TVM on Samsung Galaxy S8 (GPU: Adreno 540).}
    \vspace{0.2em}
    \begin{tabular}{|c|c|c|}
        \hline
        \#Trial & Auto-tuning & Compiling \\
        \hline
        \hline
        $1$   & $355$  & $40$ \\
        $10$  & $1477$ & $41$ \\
        $30$  & $4583$ & $41$ \\
        \hline
    \end{tabular}
    \label{tab:tvm_time_cost}
    \vspace{-1em}
\end{table}

\begin{table}[t]
    \centering
    \caption{Top-$5$ popular devices and their average inference time (AIT, ms) using MNN in a very large-scale real production case.}
    \vspace{0.2em}
    \begin{tabular}{|c|c|c|c|}
        \hline
        Device & CPU type & GPU type & AIT \\
        \hline \hline
        EML-AL00 & Kirin 970 & Mali-G72 MP12 & $87.9$  \\
        PBEM00 & SDM670 & Adreno 615 & $84.5$  \\
        PACM00 & Cortex-A73 & Mali-G72 MP3 & $92.0$ \\
        COL-AL10 & Cortex-A73 & Mali-G72 MP12 & $95.1$ \\
        OPPO R11 & Kryo 260 & Adreno 512 & $91.4$ \\
        \hline
    \end{tabular}
    \label{tab:online_time_cost}
    \vspace{-1em}
\end{table}

\subsection{Online Case Study}
Searching commodities is an indispensable stage for shopping on E-commerce platforms. However, due to the complexity of commodity categories, the traditional way of text search cannot meet the expectation of users these days. Therefore, searching commodity from images become a must-have feature for E-commerce platforms. This part shows a real application case of MNN in this scenario.

In the E-commerce application, MNN is employed to run deep learning models on mobile devices for main object detection and  detected results are then used in commodity search. This service covers more than $500$ kinds of mobile devices and has over $10$ million daily users. Table~\ref{tab:online_time_cost} shows the top-$5$ popular devices used in this service and the average inference time, where MNN achieves stable and smooth search experience with average inference time $90.2$ (ms) across all different devices, regardless of their broad diversity. This shows the encouraging universality of MNN.

\section{Conclusion and Future Work}
Mobile inference engine is critical to deep learning model deployment on mobile applications. To deal with the challenge of model compatibility and device diversity, we introduce Mobile Neural Network (MNN), which proposes a novel mobile engine design paradigm (semi-automated search) to get the best of both universality and efficiency. Pre-inference mechanism with a thorough kernel optimization and backend abstraction endows MNN favorable universality and state-of-the-art on-device inference performance.

MNN is still fast evolving, which is being improved in many aspects, for example, (1) applying auto-tuning during backend evaluation, (2) integrating model compression tools (\eg, pruning) to slim the model on the fly, (3) providing more tools for user convenience, (4) providing more language support including JavaScript and Python.

\section*{Acknowledgements}
We thank Chaoyue Niu for helpful discussions and the anonymous reviewers for their valuable comments to improve our work.

\bibliography{reference}
\bibliographystyle{sysml2020}


\appendix
\section{MLPerf Evaluation}
We also conduct MobileNet-v2 benchmark with MNN using the benchmark tool MLPerf~\cite{reddi2019mlperf} on $4$ CPU threads of Pixel 3. Results are shown in Table~\ref{tab:mlperf_results}.
\begin{table}[h]
    \centering
    \caption{MLPerf benchmark results.}
    \vspace{0.2em}
    \begin{tabular}{|c|c|}
        \hline
        Item of evaluation & Value \\
        \hline \hline
        min\_query\_count & $1024$ \\
        max\_query\_count & $5000$ \\
        QPS w/ loadgen overhead  & $64.22$ \\
        QPS w/o loadgen overhead & $64.27$ \\
        Min latency (ns) & $13212918$ \\
        Max latency (ns) & $36022504$ \\
        Mean latency (ns) & $15560004$ \\
		$50.00$ percentile latency (ns) & $15600783$ \\
		$90.00$ percentile latency (ns) & $16407241$ \\
		\hline
    \end{tabular}
    \label{tab:mlperf_results}
\end{table}

\section{More comparison on Pixel phones}
To further compare with TF-Lite, we also conduct evaluations of Inception-v3 float model on the CPU of Pixel 2 and 3. Results are shown in Table~\ref{tab:pixel_tf_lite_results}. As we see, MNN is consistently faster than TF-Lite with either single thread or multi-threads, in line with the results in the main paper.
\begin{table}[h]
    \centering
    \caption{CPU inference time (ms) comparison on Pixel phones.}
    \vspace{0.2em}
    \begin{tabular}{|c|c|c|c|}
        \hline
        Phone type & \#Threads & TF-Lite & MNN \\
        \hline \hline
        Pixel 2 & $1$ & $974$ & $\mathbf{664}$ \\
        Pixel 2 & $4$ & $310$ & $\mathbf{214}$ \\
        Pixel 3 & $1$ & $873$ & $\mathbf{593}$ \\
        Pixel 3 & $4$ & $239$ & $\mathbf{160}$ \\
		\hline
    \end{tabular}
    \label{tab:pixel_tf_lite_results}
\end{table}

\section{Backend cost evaluation}
Both CPU and GPU use \texttt{FLOPS} to measure the capability of the processors. Only GPU has the $t_{\text{schedule}}$ term. Their values are determined as follows.
\begin{itemize}
    \item \texttt{FLOPS}. For CPU, if the OS is Linux or Android, we can access the maximal frequency of each CPU core. Then choose the largest $k$ frequencies and add them together as the \texttt{FLOPS} term, where $k$ is the pre-specified number of threads (such as two threads or four threads). For the other CPU systems, set \texttt{FLOPS} to $2 \times 10^9$. For GPU, we estimate the \texttt{FLOPS} through practical running. Specifically, we run the MobileNet-v1 network for $100$ times and obtain the \texttt{FLOPS} values for a bunch of common mobile GPUs. The results are shown in the list below. For those GPUs not in this list, we set the \texttt{FLOPS} as $4 \times 10^9$, namely, faster than CPU, as is the normal case.
    
The list of GPU \texttt{FLOPS} ($10^9$): Mali-T860: $6.83$; Mali-T880: $6.83$; Mali-G51: $6.83$; Mali-G52: $6.83$; Mali-G71: $31.61$; Mali-G72: $31.61$; Mali-G76: $31.61$; Adreno (TM) 505: $3.19$; Adreno (TM) 506: $4.74$; Adreno (TM) 512: $14.23$; Adreno (TM) 530: $25.40$; Adreno (TM) 540: $42.74$; Adreno (TM) 615: $16.77$; Adreno (TM) 616: $18.77$; Adreno (TM) 618: $18.77$; Adreno (TM) 630: $42.74$; Adreno (TM) 640: $42.74$.

    \item $t_{\text{schedule}}$. This term depends on the adopted graphical API. For OpenCL and OpenGL, it is empirically set to $0.05$ (ms), which is the normal average time of calling API like {\verb+clEnqueueNDRKernel+}. For Vulkan, since it only needs to summit {\verb+commandBuffer+}, which is less time-consuming, thus $t_{\text{schedule}}$ can be estimated as $0.01$~(ms).
\end{itemize}


\end{document}